\RequirePackage{fix-cm}
\documentclass{svjour3}                     % onecolumn (standard format)
\smartqed  % flush right qed marks, e.g. at end of proof
\usepackage{graphicx}
\usepackage{natbib}

\usepackage{mathptmx}
\usepackage[english]{babel}
\usepackage[utf8x]{inputenc}
\usepackage{amsmath}
\usepackage{amssymb}
\usepackage[colorinlistoftodos]{todonotes}
\usepackage{rotating}
\usepackage{tikz}
\usepackage{subfigure,url}
\DeclareGraphicsExtensions{.pdf,.png,.jpg}

\def\dom{{\rm dom}}
\def\tvd{\sigma}

\begin{document}

\title{Understanding Concept Drift}
%\subtitle{}

\author{Geoffrey I.\ Webb, Loong Kuan Lee, Bart Goethals, Fran\c{c}ois Petitjean}

\institute{Geoffrey I.\ Webb \at
              Faculty of Information Technology, Monash University, Clayton, Vic, 3800, Australia \\
              \email{geoff.webb@monash.edu}           %  \\
           \and
            Loong Kuan Lee  \at
              Faculty of Information Technology, Monash University, Clayton, Vic, 3800, Australia \\
              \email{lklee9@student.monash.edu}           %  \\
           \and
            Fran\c{c}ois Petitjean \at
              Faculty of Information Technology, Monash University, Clayton, Vic, 3800, Australia \\
              \email{francois.petitjean@monash.edu}           %  \\
           \and
           Bart Goethals \at
              Department of Mathematics and Computer Science, University of Antwerp, Belgium, and \\
              Faculty of Information Technology, Monash University, Clayton, Vic, 3800, Australia \\
              \email{goethals@gmail.com}
}

\date{Received: date / Accepted: date}

\maketitle

\begin{abstract}
Concept drift is a major issue that greatly affects the accuracy and reliability of many real-world applications of machine learning.  We argue that to tackle concept drift it is important to develop the capacity to  describe and analyze it. We propose tools for this purpose, arguing for the importance of quantitative descriptions of drift in marginal distributions.  We present quantitative drift analysis techniques along with methods for communicating their results.  We demonstrate their effectiveness by application to three real-world learning tasks.
\end{abstract}

\section{Introduction}
\label{intro}

The world is dynamic, in constant flux. But machine learning usually creates static models from historical data. As the world changes, these models can grow increasingly unreliable. There has been a very substantial effort investigating methods for detecting concept drift \citep{Kifer:2004:DCD:1316689.1316707,gama2004learning,baena2006early,dries2009adaptive,bifet2013cd,qahtan2015pca}. However, in applications where drift is continual, these may be of limited use, as they should always flag drift as present. In such cases, rather than simply flagging whether drift is occurring or not, it may be more useful to generate a detailed description of the nature and form of the drift. We call such a description a \emph{concept drift map}.  

\begin{figure}
	\centering
	\includegraphics[scale=0.42,trim=15pt 2.2in 0pt 2.1in, clip]{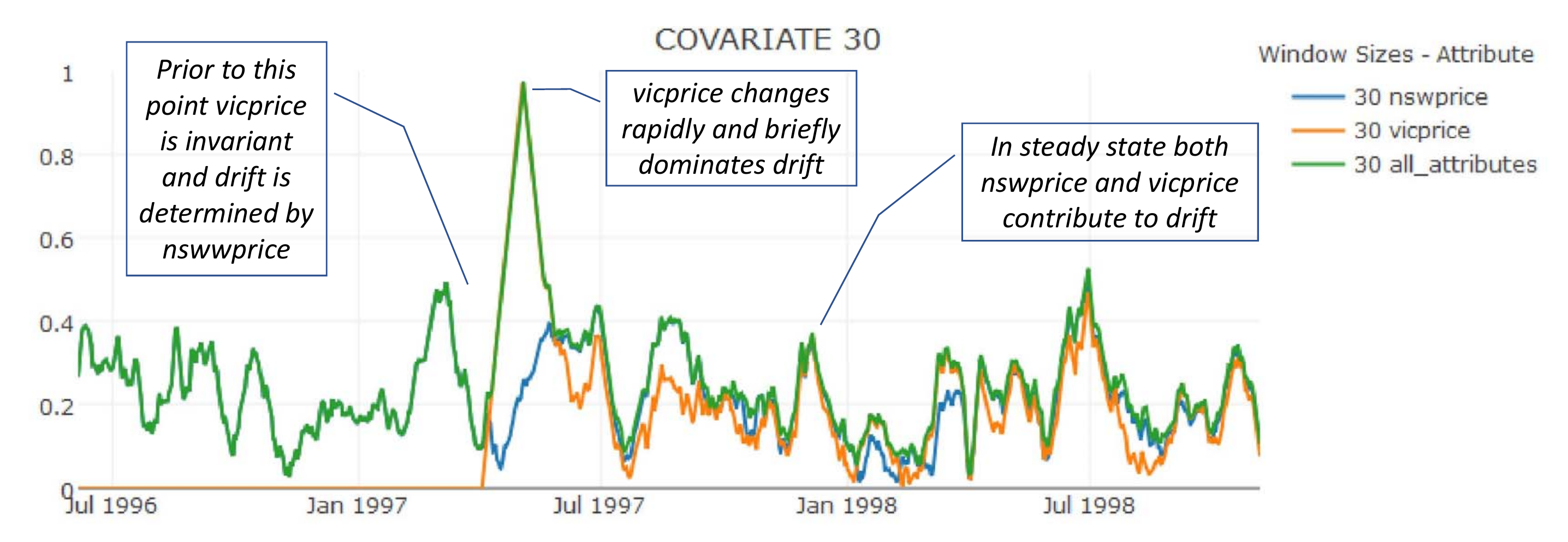}
	\caption{An example concept drift map plotting the changing drift magnitude of two variables both individually and jointly. These variables record the electricity price in each of the Australian states of Victoria and New South Wales. Prior to May 2$^{\textrm{nd}}$ 1997 there was no interstate electricity market and the Victorian price was invariant.}
	\label{fig:example}
\end{figure}

Figure \ref{fig:example} shows a simple map for the electricity data, explained in detail in Section \ref{sec:eletricity}. In this simple example concept drift map we show the drift in two key variables \texttt{nswprice} and \texttt{vicprice} both individually and jointly. This map shows how \texttt{nswprice} determines the drift up until May 1997, then \texttt{vicprice} briefly dominates before the electricity market settles and each contributes to the joint drift.

We envisage many applications for these maps. They might be used by online learning algorithms to manage response to drift. They could be employed to understand the form of drift affecting an application in order to assist in selecting strategies to deal with it. They could also help to better understand the relative performance of alternative drift mechanisms on different tasks in order to improve understanding of the relative capabilities of the current alternatives.%\todo{GW: Are there other applications?}

In this paper we present techniques for generating drift maps from  data. In Section~\ref{sec:description} we provide a formal definition of the problem and related terminology. In Section~\ref{sec:total-magnitude} we present methods for measuring total drift magnitude. In Section~\ref{sec:marginal-magnitude} we present methods for measuring marginal drift magnitudes.
Section~\ref{sec:communication} describes graphical methods for communicating the detailed maps that our quantitative techniques produce.
Section~\ref{sec:examples} evaluates the effectiveness of our techniques on three real-world datasets.
We present conclusions in Section~\ref{sec:conclusion}.

\section{Problem description}\label{sec:description}

A data stream is a data set in which the objects have time stamps, which, depending on the granularity of the stamps, induces either a total or a partial order between observations. 
While our techniques generalize in a straightforward manner to situations in which there is no target attribute, in the current work we assume a classification learning context. In~consequence, we can consider the process that generates the stream to be a joint distribution over random variables $Y$ and $X=\{X_1,\ldots,X_n\}$, where $y\in \dom(Y)$ are the class labels and the $x_i\in \dom(X_i)$ are the attribute values.  
We provide a summary of the key symbols used in this paper in Table~\ref{tab:symbols}.

\begin{table}[t]
	\begin{center}
		\begin{tabular}{|l|p{2.65in}|l|}
			\hline
			\textbf{Symbols}& 
			\textbf{Represents}& 
			\textbf{Scope} \\
			\hline
			{$t, u, v, w$}& 
			Points of time& 
			$\mathbb{R}_{{>}0}$ \\
			\hline
			{$m$}& 
			Durations& 
			$\mathbb{R}_{{\geq}0}$ \\
			\hline
			{$i, j, k, n$}& 
			General purpose non-negative integers& 
			$\mathbb{N}$ \\
			\hline
			$X=\{X_1,\ldots,X_n\}$&
			A random variable over covariates& 
			Stream dependent \\
			\hline
			{$x_i$}&
			A value of a covariate& 
			$\dom(X_i)$ \\
            \hline
            $\bar{x}=\langle x_1,\ldots,x_n\rangle$ &
            A simultaneous assignment of values to all covariates &
            $\dom{X}$\\
			\hline
			$Y$&
			A random variable over class labels& 
			Stream dependent \\
			\hline
			$y$&
			A class label& 
			$\dom(Y)$ \\
			\hline
			$Z$&
			A vector of random variables over either covariate values or class labels& 
			Stream dependent \\
			\hline
			$\bar{z}i$&
			A value in $\dom(Z_i)$& 
			$\dom(Z_i)$ \\
			\hline
            $P_{t}\left(X, Y\right)$&
            A probability distribution at time $t$&
            \\
			\hline
            $P_{[t,u]}\left(X, Y\right)$&
            A probability distribution over time period $[t,u]$&
            \\\hline
            $\tvd_{t,u}(Z)$&Total variation distance measure of drift between each of the probability distributions of vector of random variables $Z$ from time period $t$ to $u$&[0,1]\\\hline
            $\sigma^{X\mid Y}_{t,u}$&Conditioned marginal covariate drift between each of the probability distributions from time period $t$ to $u$&[0,1]\\\hline
$\sigma^{Y\mid X}_{t,u}$&Posterior drift between each of the probability distributions from time period $t$ to $u$&[0,1]\\\hline
		\end{tabular}
		
		% \comment{Check that we actually use all of these symbols - I suspect we just use a for concepts, for example.  We should also include all of the greek letters. }
		
		\caption{List of symbols used.}
		\label{tab:symbols}
	\end{center}
\end{table}

%We use $P(Y)$ to denote the prior probability distribution over the class labels, $P(X)$ to denote the prior probability distribution over covariates, $P(X,Y)$  to denote the joint probability distribution over objects and class labels, $P(Y\mid X)$ to denote the likelihood distribution over class labels given objects and $P(X\mid Y)$ to denote the posterior probability distribution over covariates given class labels.  

In order to reference the probability distribution at a particular time we add a time subscript, such as $P_t(X,Y)$, to denote a probability distribution at time $t$. It is often not practical to estimate the distribution in effect at a specific point in time and for this purpose we often deal with concepts and probability distributions over a time interval,  $P_{[t,u]}(X,Y)$.

We follow recent practice and adopt \citeauthor{SurveyConceptDriftAdaptation}'s (\citeyear{SurveyConceptDriftAdaptation}) definition of a concept.
\begin{equation}
	\textrm{Concept} = P\left(X, Y\right).
\end{equation}
In the context of a data stream, we need to recognize that concepts may change over time. To this end we  define the concept at a particular time $t$ as  
\begin{equation} \label{eq:ConceptTime}
	P_{t}\left(X, Y\right)
\end{equation}
and at a specific time period $[t,u]$ as
\begin{equation} \label{eq:ConceptTimePeriod}
	P_{[t,u]}\left(X, Y\right).
\end{equation}

Concept drift occurs between times $t$ and $u$ when the distributions change, 
\begin{equation}\label{eq:drift-definition}
	P_t(X, Y)\neq P_u(X, Y)
\end{equation}
and similarly between time periods $[t,u]$ and $[v,w]$,
\begin{equation}\label{eq:drift-definition-periods}
	P_{[t,u]}(X, Y)\neq P_{[v,w]}(X, Y).
\end{equation}

We define the \emph{concept drift mapping task} as taking as input a data stream and generating as output useful descriptions of the drift in the process that generates the data.

\section{Measuring total drift magnitude}\label{sec:total-magnitude}

\cite{webb2015characterizing} proposed four quantitative measures of concept drift including the key measure \emph{drift magnitude} which measures the distance between two concepts $P_t(X, Y)$ and $P_u(X, Y)$. Any measure of distance between distributions could be employed. \cite{webb2015characterizing} use Hellinger Distance \citep{HellingerDistance} for this purpose.  In the current work we employ Total Variation Distance \citep{levinmarkov}:
\begin{equation}\label{eq:TVD}
\tvd_{t,u}(Z)=\frac{1}{2}\sum_{\bar{z}\in\dom(Z)}
\left|P_t(\bar{z})-P_u(\bar{z})\right|
\end{equation}
where $Z$ represents a vector of random variables.
\iffalse\todo[inline]{Not sure this makes sense. A random variable maps to a numerical outcome, in our case I think that $Z$ can map to states that are not ordered (eg strings). Moreover, this "random varialbe over any set of" is confusing. What about "where $Z$ can represent any set of variables". (but then what does dom(Z) means?)}
\todo[inline,color=pink]{GW: According to no less authoritative source than Wikipedia ``Part of the definition of a random variable is the set of possible outcomes. In the case of the coin, there are only two possible outcomes, namely heads or tails.'' So I don't think a random variable is restricted to ordinal outcomes.  How about the following?}
\begin{equation}
\tvd_{t,u}(Z_1,\ldots,Z_m)=\frac{1}{2}\sum_{{z_1}\in\dom(Z_1),\ldots,{z_m}\in\dom(Z_m)}
\left|P_t(z_1,\ldots,z_m)-P_u(z_1,\ldots,z_m)\right|.
\end{equation}
\fi

Of all the standard measures of distance between probability distributions we favour Hellinger Distance and Total Variation Distance because they are metrics and it seems highly desirable that a measure of drift between two periods should be symmetric.
In this paper we use Total Variation Distance because it
it is slightly less complex to analyse and more efficient to compute, but our approaches trivially generalize to Hellinger Distance.

Note that we limit our techniques to discrete valued data. While there are techniques for computing Total Variation and Hellinger Distance for continuous data drawn from specific distributions, such as a Gaussian, these require strong assumptions about the form of the distribution. Further, handling of numeric and categorical data together adds additional complexities. In consequence, we discretize all numeric attributes, using 5 bin equal frequency discretization of each attribute across all time periods.

\cite{webb2015characterizing} propose a number of quantitative measures for drift that provide gross summaries of the drift between two time points.  These include using any measure of distance between probability distributions to measure drift magnitude. They demonstrated that these measures enable insights to be derived that are otherwise not possible, such as how different algorithms perform in the face of drift of varying magnitude.  However, our subsequent uses of these measures in real world applications have revealed that it can be important to augment these overview measures with further finer grained analysis.

One limitation of a single gross measure of drift magnitude arises from both Total Variation Distance and Hellinger Distance being monotonic as the dimensionality of data increases. We provide a proof of this in Appendix~\ref{appendix:monotonicity}. As a result, in practice, in high dimensional data these measures are likely to be close to their maximum, 1.0, simply through accumulation of small differences across many dimensions.  This reduces their capacity to distinguish between cases of drift.

Second, a single gross measure of drift fails to recognize or to describe the details of how drift differs across the subspaces defined on different attributes of the data. In the real world, drift is often not uniform, as we show in Section \ref{sec:examples}.  For example, not all factors are subject to inflation and those that are may increase at varying rates. A change in technology may cause a sudden abrupt change in some  attributes of the data but have no affect whatsoever on others. Some factors may drift in cycles with differing periodicity and other factors may be subject to drift that is not cyclical. In many real world applications it is likely to be useful to be able to understand which attributes and combinations of attributes are drifting in which manners at any particular time.

For these reasons we investigate the introduction of \emph{concept drift maps}, methods for describing the drift affecting different subspaces of the data.

\section{Measuring marginal drift magnitude}\label{sec:marginal-magnitude}

The key to describing drift in different attribute subspaces is to measure the drift in the marginal distributions defined over different combinations of attributes.

A problem that arises is how to estimate the required probability distributions from the available data. In order to manage the variance in the estimates it is important to derive them from sufficiently large data samples.  This will usually preclude the possibility of deriving instantaneous estimates --- estimates of the probability distribution at any single point in time.  Rather it will often be necessary to derive estimates of the distribution over some time interval, such as the distribution for a given hour, day or week. However, this practical driver is not the only reason for considering drift between extended periods rather than drift between instantaneous points in time. As we show in Section \ref{sec:examples}, consideration of drift between periods of differing granularity can also be extremely revealing. In consequence, our techniques estimate the drift between two time intervals by first estimating the distributions in each interval and then calculating the magnitude of the drift between them.

In the current work we use maximum likelihood estimates.

It turns out to be useful to map not only the drift in the \emph{joint distribution} $P(X,Y)$, but also the \emph{covariate distribution} $P(X)$, the \emph{class distribution} $P(Y)$, the \emph{posterior distribution} $P(Y\mid X)$ and the \emph{conditioned covariate distribution} $P(X\mid Y)$, as each reveals different facets of a potentially complex drift.  For joint, covariate and class distributions Equation \ref{eq:TVD} applies directly.  However, for the two conditional distributions it is necessary to deal with multiple distributions, one for each value of the conditioning attributes. We address this by weighted averaging, as described in the next two subsections..

\subsection{Conditioned Marginal Covariate Drift.}
For a given subset of the covariate attributes there will be a conditional probability distribution over the possible values of the covariate attributes for each specific class, $y$. 
The \emph{conditioned marginal covariate drift} is the weighted sum of the distances between each of these probability distributions from time period $t$ to $u$, 
where the weights are the average probability of the class over the two time periods.
\begin{equation}
    \sigma^{X\mid Y}_{t,u} = \sum_{y \in Y} \left[\frac{P_t(y) + P_u(y)}{2}  \frac{1}{2} \sum_{\bar{x} \in X} |P_t(\bar{x}\mid y) - P_u(\bar{x}\mid y)|\right]
\end{equation}

\subsection{Posterior Drift.}
For each subset of the covariate attributes there will be a probability distribution over the class labels for each combination of covariate values, $\bar{x}$ at each time period. 
Therefore, the Posterior Drift can be calculated as the weighted sum of the distances between these probability distributions 
where the weights are the average probability over the two periods of the specific value for the covariate attribute subset.
\begin{equation}
    \sigma^{Y\mid X}_{t,u} = \sum_{\bar{x} \in X} \left[\frac{P_t(\bar{x}) + P_u(\bar{x})}{2} \frac{1}{2} \sum_{y \in Y} |P_t(y\mid \bar{x}) - P_u(y\mid \bar{x})|\right]
\end{equation}

\section{Methods for communicating drift maps}\label{sec:communication}

Our primary technique measures marginal drift magnitudes between time periods. Sometimes it will be interesting to consider a single such comparison at a time.  At other times it will be useful to consider how drift unfolds over an extended period of time.  This can result in very large numbers of individual drift values. Here we present methods for succinctly communicating these large amounts of information.

For drift over the marginals between two time periods the key  information that we want to convey is the relative magnitude of the drift in each combination of attributes. We find that heat maps provide a highly effective means of doing so, clearly highlighting the interactions between the variables. We provide examples in Figures \ref{fig:Prior2} to \ref{fig:Pos2} below.

We use line plots to communicate the evolution of drift over extended periods of time. In doing so we use two  periodicity parameters. The first parameter is how frequently should the drift be calculated.  In the Electricity and Airlines domains discussed below, we calculate the drift daily.  The second parameter is the period over which to determine the distributions to be compared. In the Airlines domain we use two periods for this purpose, daily and weekly, and show that each reveals different insights.

%The total variation distance is a sum over all of the combinations of attribute values. In some cases further insight can be derived by considering the relative contribution to this sum of each attribute value combination.  We provide examples of this in Section \ref{sec:eletricity}.

\section{Illustrative examples}\label{sec:examples}
We illustrate the proposed techniques by application to a number of real-world datasets. 

\subsection{Electricity}\label{sec:eletricity}
The first example is electricity pricing in South-East Australia, a dataset downloaded from the MOA dataset repository \citep{MOAdatasets} and described by \cite{Harries99splice-2comparative}.  The covariates are \texttt{nswprice}, \texttt{nswdemand}, \texttt{vicprice}, \texttt{vicdemand}, and \texttt{transfer}, recording the price and demand in the states of New South Wales and Victoria and the amount of power transferred between the states. The class label identifies whether the transfer price is increased or decreased relative to a moving average of the last 24 hours. Examples are generated for every 30 minute period from 7 May 1996 to 5 December 1998. The values have been normalized to the interval [0,1].

\begin{figure}
	\centering
	\includegraphics[scale=0.54,trim=25pt 20pt 0pt 0pt, clip]{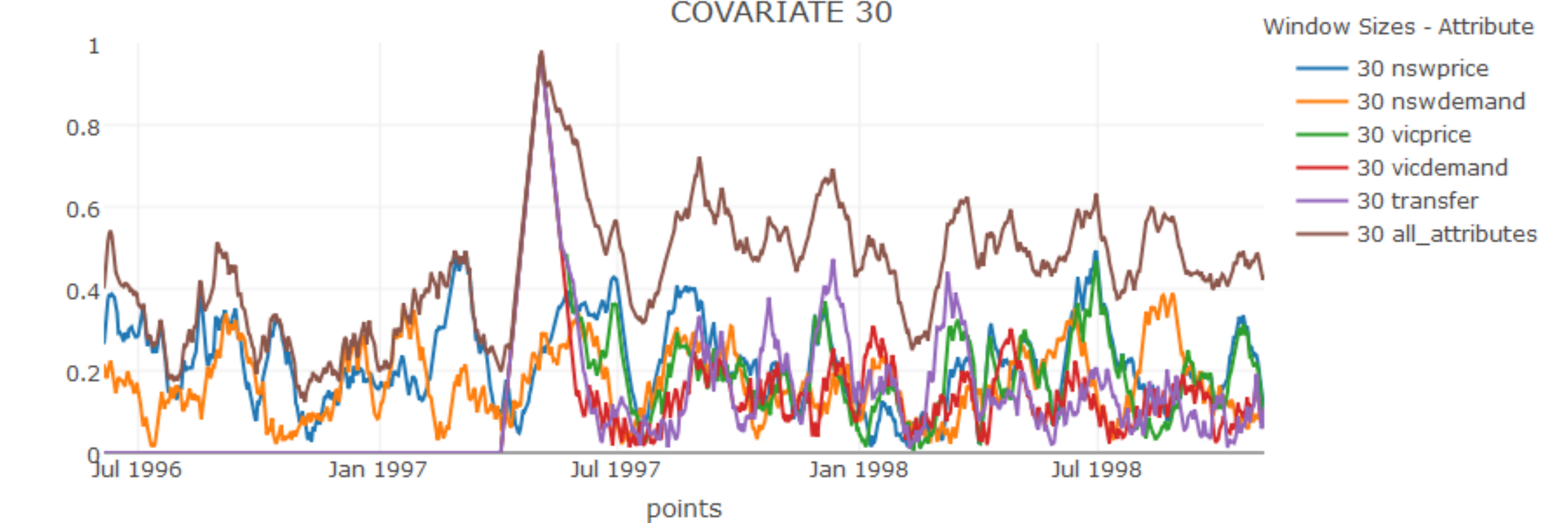}
	\caption{Covariate drift for the electricity data. Values calculated daily for the drift between the 30 days prior to  the current day and the 30 days thereafter.}
	\label{fig:elec-covariate}
\end{figure}%
\begin{figure}
	\centering
	\includegraphics[scale=0.54,trim=25pt 20pt 0pt 0pt, clip]{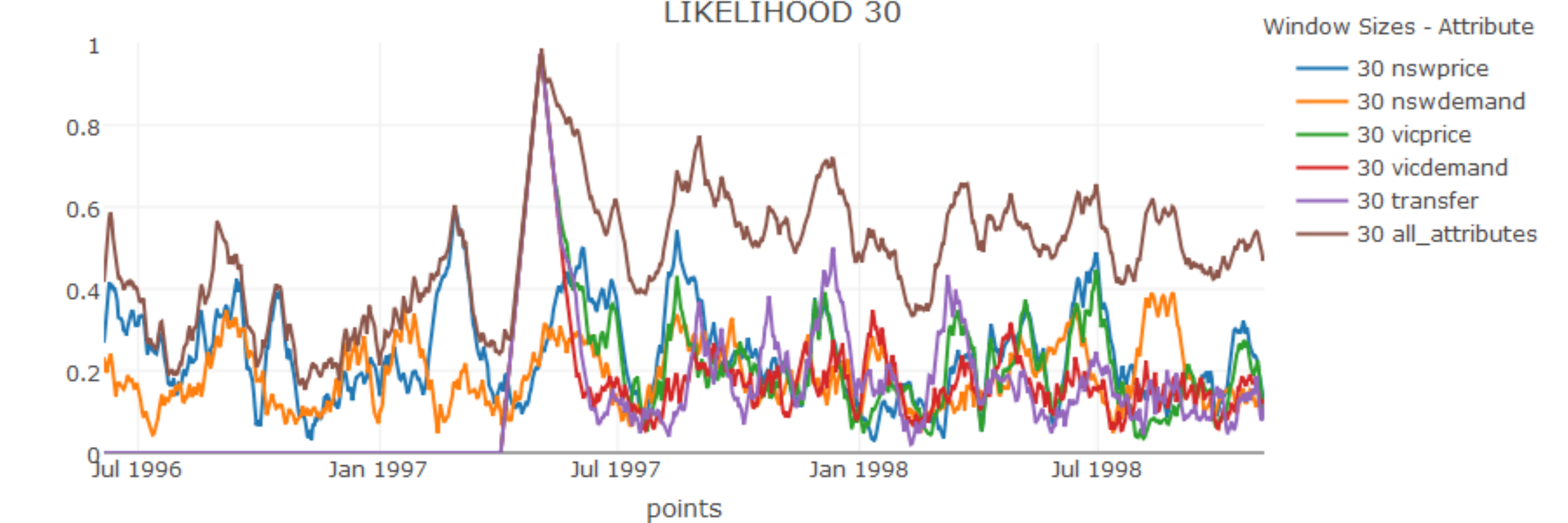}
	\caption{Conditioned covariate drift for the electricity data. Values are calculated daily for the drift between the 30 days prior to the current day and the 30 days thereafter.}
	\label{fig:elec-condcovariate}
\end{figure}%
\begin{figure}
	\centering
	\includegraphics[scale=0.54,trim=25pt 20pt 0pt 0pt, clip]{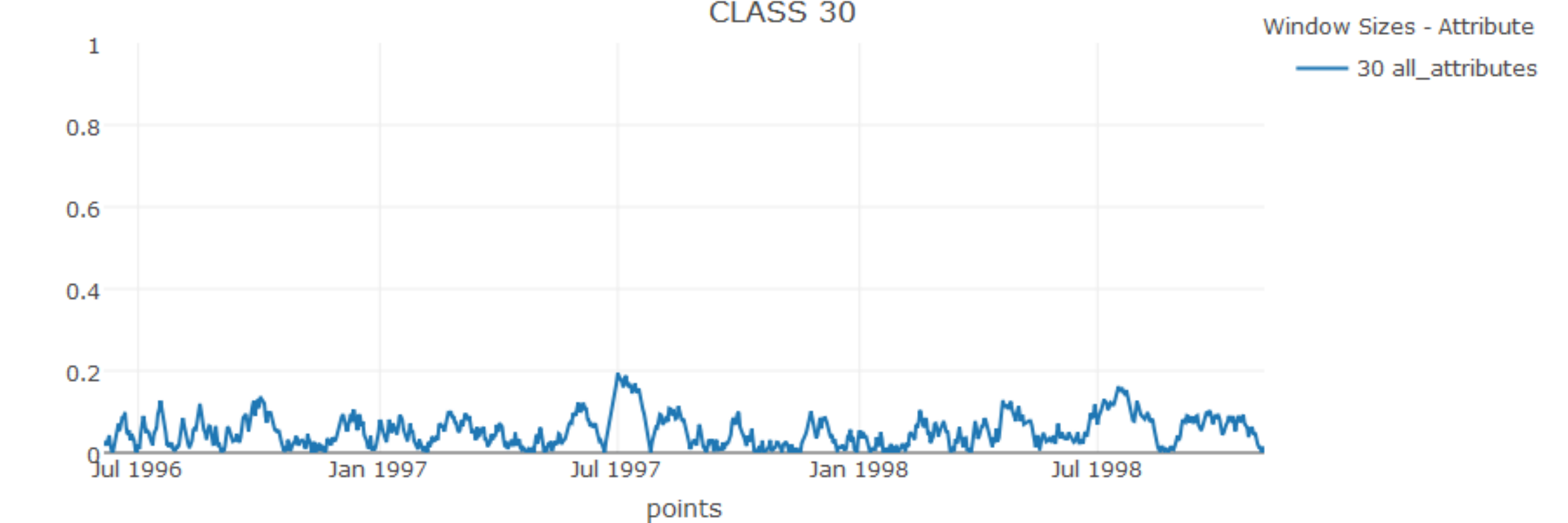}
	\caption{Class drift for the electricity data. Values are calculated daily for the drift between the 30 days prior to the current day and the 30 days thereafter.}
	\label{fig:elec-class}
\end{figure}%
Figures \ref{fig:elec-covariate} and \ref{fig:elec-condcovariate} present the covariate drift and conditioned covariate drift respectively.
Each point corresponds to a day and presents the drift from the 30 day period prior to that day compared to next 30 days.  

As can be seen, there is a sudden increase in covariate drift on 2$^{\textrm{nd}}$ of May 1997. This is the date at which the process of introducing a national electricity market (NEM) commenced. From this date a trial NEM allowed wholesale electricity sales  between the states of New South Wales, Victoria, the Australian Capital Territory, and South Australia \citep{Roarty1998Electricity}.  
\texttt{Vicprice}, \texttt{vicdemand} and \texttt{transfer} have no drift prior to this date. Indeed these three variables are constant until the market is introduced. Past May 1997, drift in \texttt{nswprice} and \texttt{nswdemand} stays similar to before, but substantial variability is apparent in the drift within Vic\texttt{vicprice}, \texttt{vicdemand}, and \texttt{transfer}. The conditioned covariate drift closely follows the unconditioned covariate drift indicating that there was little difference in drift of the covariates between classes.  This illustrates how our proposed mapping of drift over both marginals as well as the distribution as a whole can provide additional useful information. 

The class drift shows relatively low levels of drift. Note that as this is a binary variable, high levels of drift in this map would indicate that the transfer price has trended in one direction (up or down) for the previous 30 days and then in the opposite direction for the following 30 days. This plot indicates that there were no such extended changes.

To summarize this example, drift increases substantially after May 2$^{\textrm{nd}}$ 1997.  The increase in drift is dominated by covariate drift and the covariate drift is dominated by drift in 3 of the five covariates, VicPrice, VicDemand and Transfer.

\subsection{Airlines}
\begin{figure}
	%\centering
	\includegraphics[scale=0.53,trim=25pt 20pt 0pt 0pt, clip]{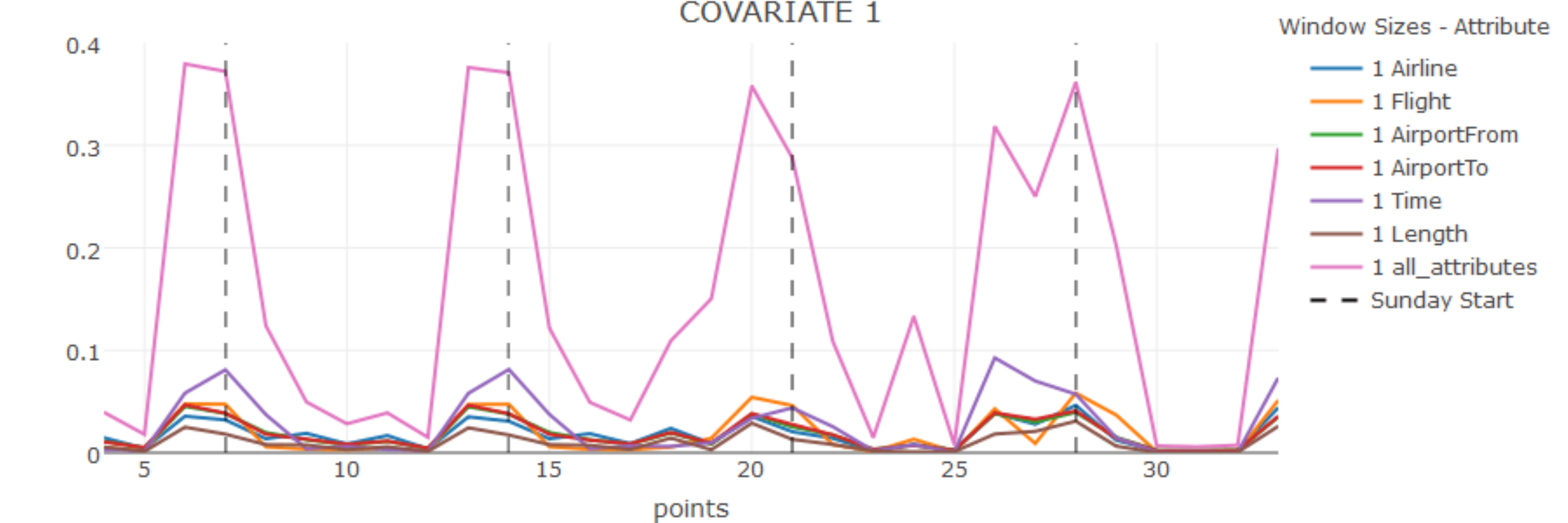}
	\caption{Daily covariate drift for the airlines data. The dashed lines are placed between each Saturday and Sunday.}
	\label{fig:airline-covariate-daily}
%\end{figure}%
%\begin{figure}
\vspace{10pt plus 10pt}
%\centering
	\includegraphics[scale=0.53,trim=25pt 20pt 0pt 0pt, clip]{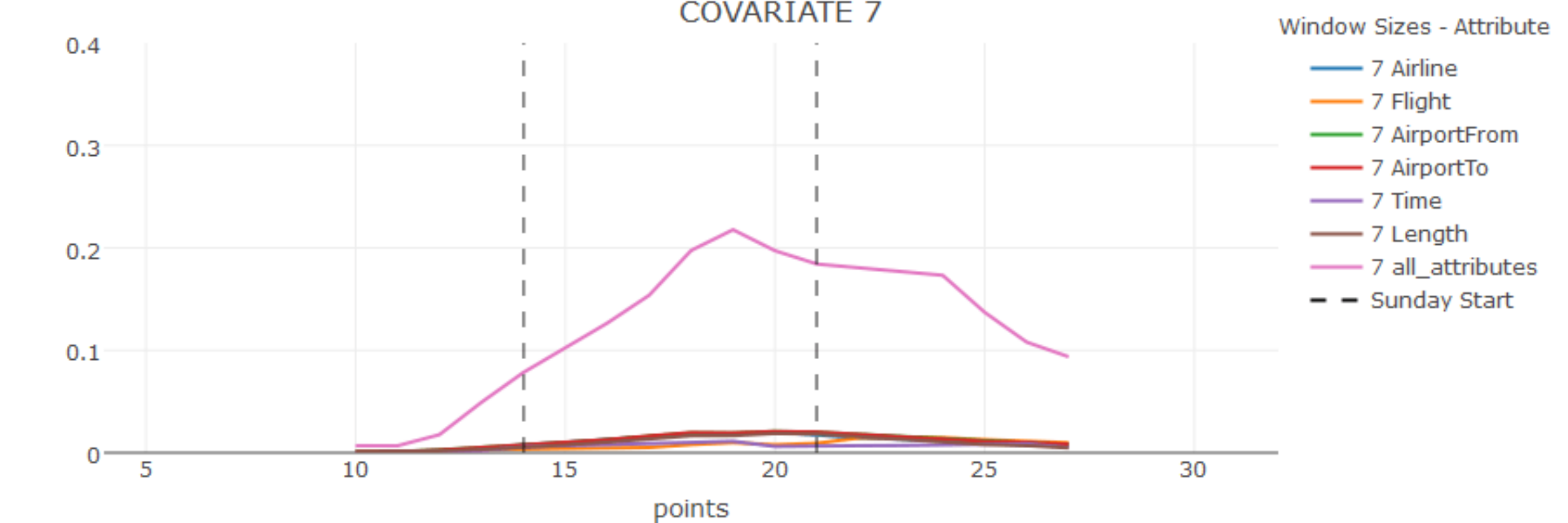}
	\caption{Weekly covariate drift calculated daily for the airlines data. The dashed lines are placed between each Saturday and Sunday.}
	\label{fig:airline-covariate-weekly}
%\end{figure}%
%\begin{figure}
\vspace{10pt plus 10pt}
%\centering

	\includegraphics[scale=0.54,trim=25pt 20pt 0pt 0pt, clip]{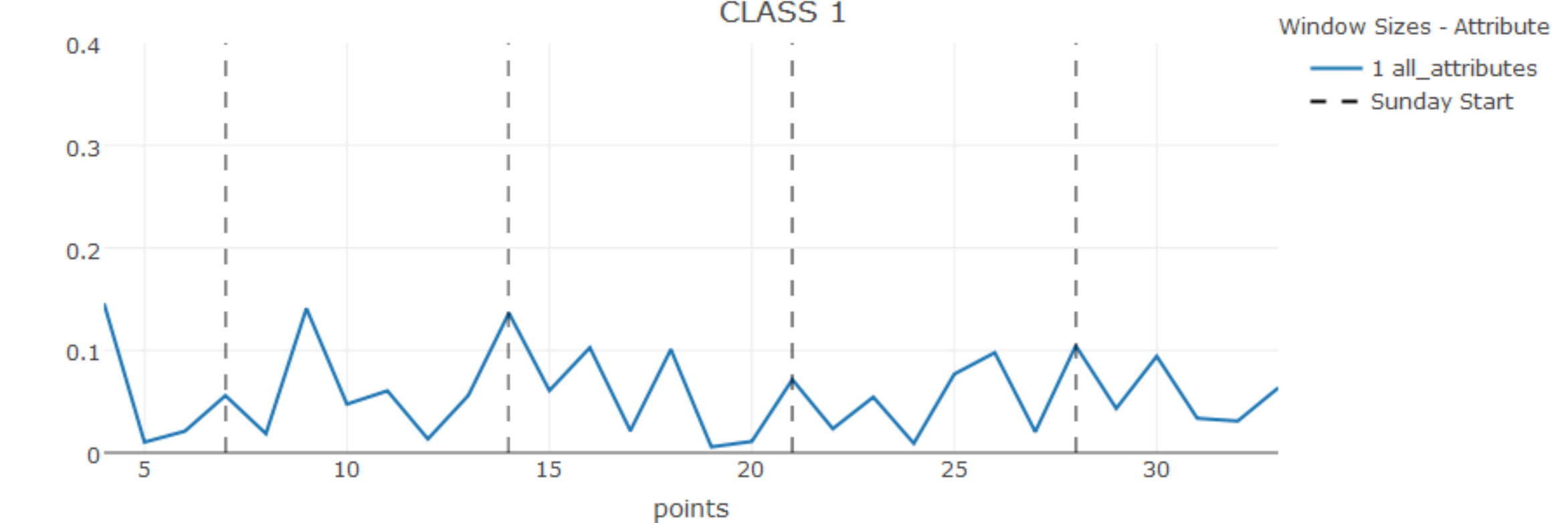}
	\caption{Daily class drift for the airlines data. The dashed lines are placed between each Saturday an Sunday.}
	\label{fig:airline-class-daily}
%\end{figure}%
%\begin{figure}
\vspace{10pt plus 10pt}
%\centering

	\includegraphics[scale=0.54,trim=25pt 20pt 0pt 0pt, clip]{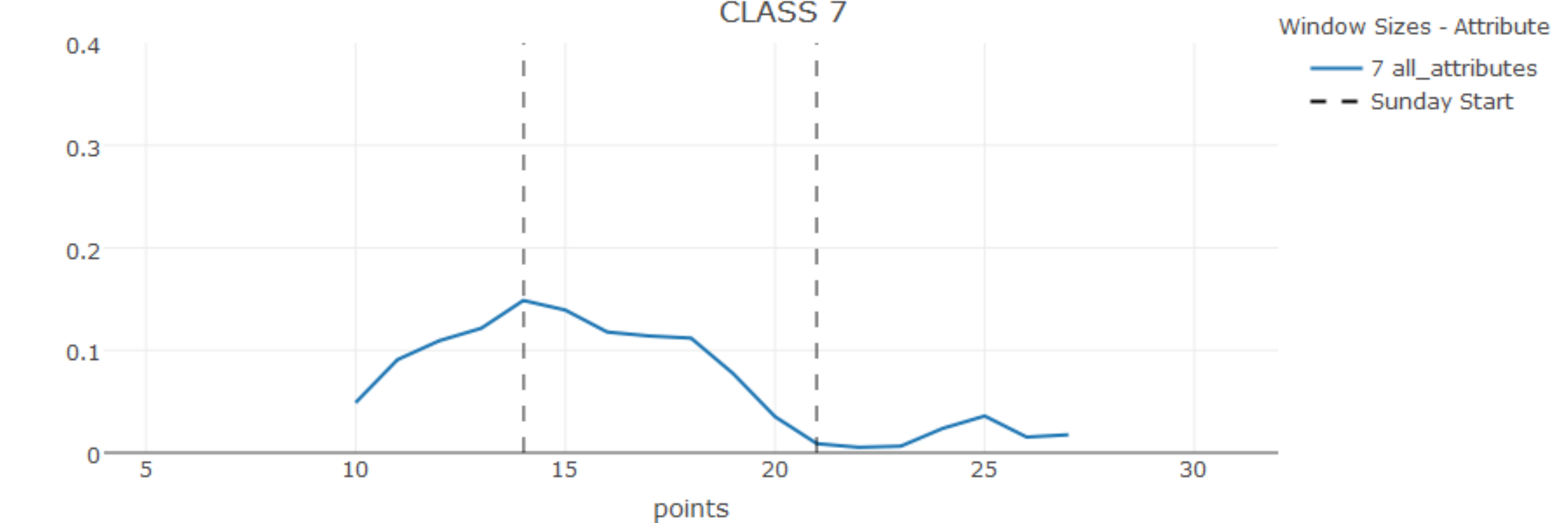}
	\caption{Weekly class drift calculated daily for the airlines data. The dashed lines are placed between each Saturday an Sunday.}
	\label{fig:airline-class-weekly}
\end{figure}%
The second example is the airlines dataset, also downloaded from the MOA dataset repository \citep{MOAdatasets}.
Each example in this data represents a flight, with covariates \texttt{Airline}, \texttt{Flight}. \texttt{AirportFrom}, \texttt{AirportTo}, \texttt{DayOfWeek}, \texttt{Time} and \texttt{Length} and with a binary class indicating whether the flight arrived on time. The \texttt{DayOfWeek} has been used to partition the data into days and weeks and have not been included as a covariate in the analysis. Figure \ref{fig:airline-covariate-daily} shows the covariate drift from day to day.  Figure \ref{fig:airline-covariate-weekly} shows the covariate drift for the week prior to a day against the week starting with that day and is plotted daily from the seventh day. Note that the numbering starts with 4 as the first day in the data is day number 3.

The first figure shows that for the first two weeks there is a cyclical pattern in the magnitude of covariate drift, with large changes from Friday to Saturday and from Saturday to Sunday, but lower drift from Sunday to Monday and substantially lower drift between successive weekdays.  However, this pattern breaks down over the following two weeks.  Unfortunately we do not have the dates for which the data were collected and hence can only speculate for the reasons for this change in pattern; weather and public holidays being two potential explanations. The marginal distributions indicate that the time of day is the major contributor to drift for most of the period but that flight number overtakes it for some parts of the second half of the period.

The weekly analysis shows that while there is substantial drift from day to day, there is little drift between the first two weeks, confirming the notion that they follow a steady cycle.  The inter-week drift then rises sharply.  Interestingly, it is the origin and destination airports and flight lengths that change most from week to week as opposed to the time of day and flight number which dominated the inter-day drift.

Figures \ref{fig:airline-class-daily} and \ref{fig:airline-class-weekly} show the daily and weekly class drift, respectively.  They reveal that the class, representing on-time performance, is not subject to the same weekly cycle of drift as the covariates and that there is greatest drift in on-time performance between the second and third weeks. It is interesting to contrast the inter-week covariate drift to the inter-week class drift.  The covariates start with almost no drift which then increases substantially, while the class starts with substantial drift and subsequently drops to having almost no drift. In~general, these plots are revealing in that they show that the class drift for this data is quite different in nature to the covariate drift.

This data demonstrates the importance of the granularity of the time periods used in drift analysis and the manner in which different granularities can each convey different and valuable insights.

\subsection{Satellite}
The final example is satellite data of land usage in France. 
We use 15 Landsat-8 images acquired over the agricultural year 2013. 
Images were obtained through the Theia Land Data Centre (\url{http://www.theia-land.fr/en/presentation/products}). The Landsat products are orthorectified prior to their release by the USGS and then, Theia processing chains based on the algorithms described in \cite{Hagolle2015} (and cloud shadow) screening and atmospheric corrections. These corrections ensure that the values observed cover the exact same geographic areas and that they are comparable over time. 

\begin{figure}
  \centering
  \includegraphics[width=.9\linewidth]{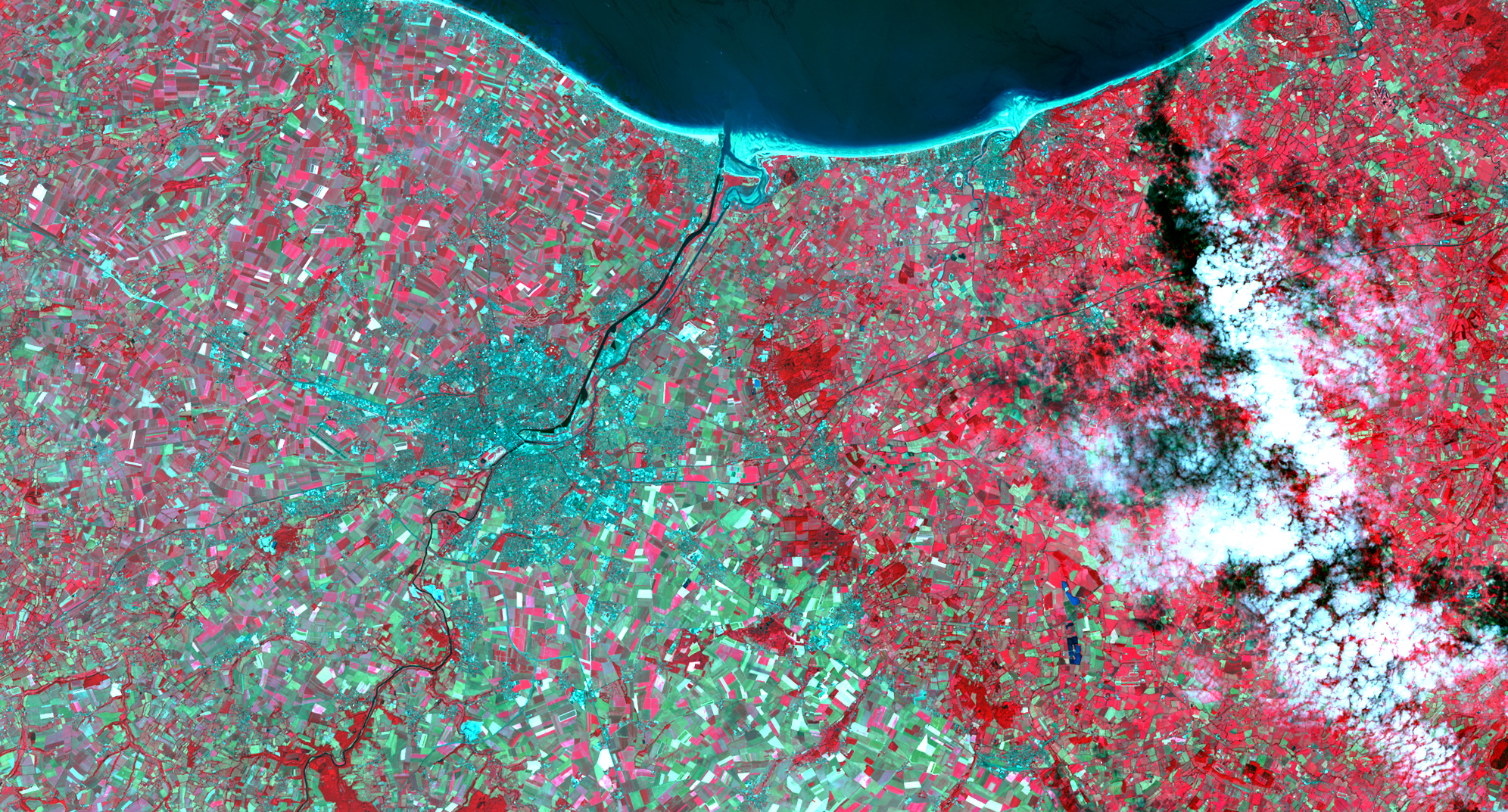}
  \caption{Landsat-8 image taken on the 17\textsuperscript{th} of July 2013 -- red displays near-infrared, green displays red and blue displays green (traditional false-color composite). Contains USGS/NASA Landsat Program data \copyright{} 2013 processed at level 2A by CNES for THEIA Land data centre. }
  \label{pic:image-france}
\end{figure}

From these images, we use the multi-spectral product at a spatial
resolution of 30~meters (Landsat-8 band 1 to band 7) and add three additional attributes, which are indices of vegetation, water and brightness (resp. Normalized Difference Vegetation Index, NDVI, Normalized Difference Water Index, NDWI, and Brightness). 
An example Landsat-8 image is illustrated in Figure~\ref{pic:image-france}.
 \citep{Inglada2017}; please see acknowledgements. 

\begin{figure}
  \centering
  \includegraphics[width=.9\linewidth]{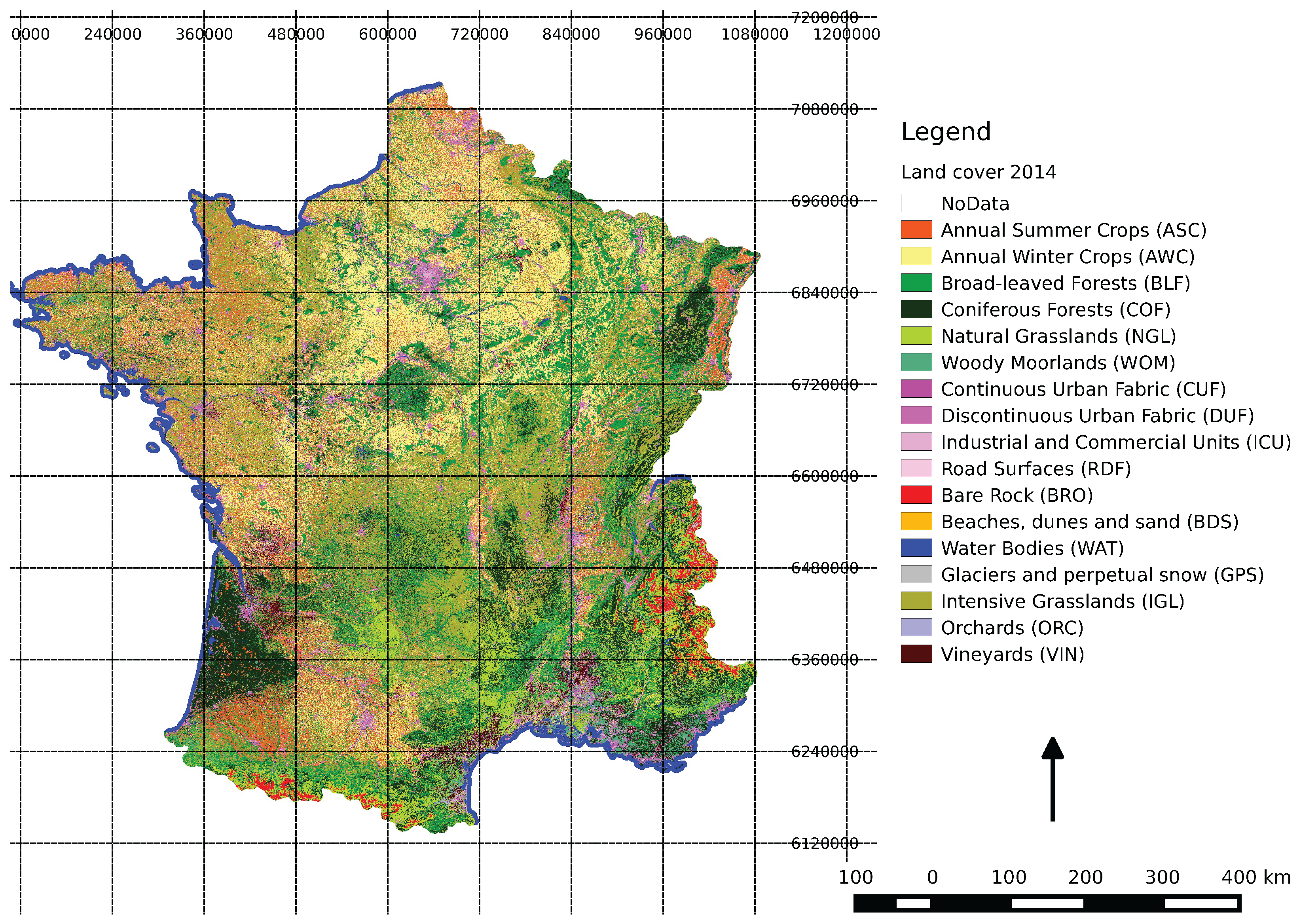}
  \caption{Labels for the Satellite dataset. \copyright{} \cite{Inglada2017} available at \texttt{http://dx.doi.org/10.3390/rs9010095} under the terms and conditions of the Creative Commons Attribution (CC BY) license (\texttt{http://creativecommons.org/licenses/by/4.0/}).}
  \label{pic:label-france}
\end{figure}

In addition, we have a land-cover map for the whole year which associates a class label to each ``pixel'' (or line) in our database; this label map is illustrated in Figure~\ref{pic:label-france}.
The data was prepared by our colleagues at the CESBIO laboratory (see acknowledgements).

The \texttt{id} class represents the land usage of the point being imaged.
We analyse here the drift between the images take on 5 May 2013 and 29 November 2013. These dates in Spring and Fall were chosen as ones between which there should be expected to be substantial changes. May is generally just before the harvest of winter crops, e.g. wheat, canola and barley (light yellow in Figure~\ref{pic:label-france}). %\todo{Could we have the trend plot as well? }

The drift magnitudes reveal that this is indeed the case.  The covariate drift magnitude is 0.68, the conditioned covariate drift is 0.76, the class drift is 0.00 and the posterior drift is 0.48. There is no class drift because the land usage is determined on an annual basis and hence does not change during the year. There is nonetheless drift in the posterior (the class conditioned on the covariates) because when $P(Y)$ is invariant and $P(X)$ changes it follows that there must be a change in $P(Y\mid X)$.

\begin{figure}
	\centering
	\includegraphics[scale=0.3]{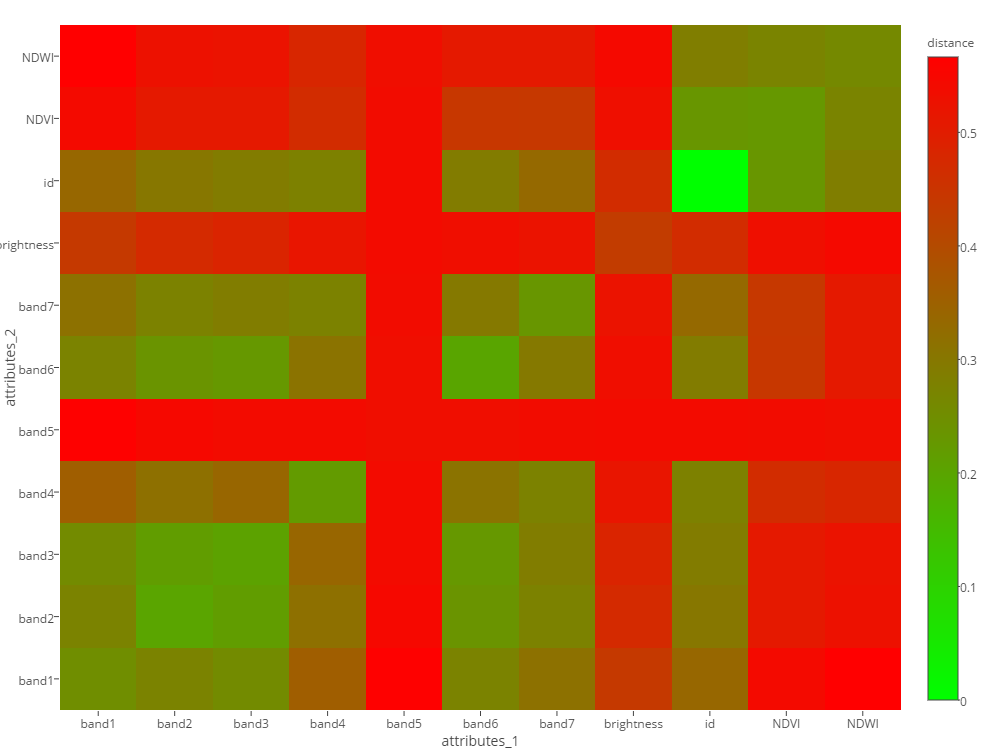}
    \caption{Pairwise drift in the joint distribution on the Satellite data between May and November}
    \label{fig:Prior2}
\end{figure}%
Figure \ref{fig:Prior2} is a heat map displaying the drift over each pair of attributes in the joint distribution on the satellite data.  The diagonal represents univariate drift. For example, the cell at the intersection of the row and column labelled \texttt{id} gives the magnitude of the drift for the class attribute \texttt{id}.  As the land usage assigned to each point does not change over the period, the drift magnitude is 0.0. The largest univariate drift is for band 5, which corresponds to near-infrared. This is explained by the fact that chlorophyll reflects near-infrared; in May, a lot of surfaces are covered by growing crops, which leads to a large amount of near-infrared being reflected. On the other hand, most crops have been harvested late November. More generally, it can be seen that each of these univariate drifts is lower than any of the bivariate drifts involving that same attribute, as our monotonicity proof in Appendix \ref{appendix:monotonicity} demonstrates they must.

The drift for \texttt{NDWI} and \texttt{NDVI} is particularly interesting.  The univariate drift for both these attributes considered in isolation is relatively low, but when considered in conjunction with most other attributes is high. %\todo[line]{Francois: I cannot explain it... Could I get the plots of the distribution of NDVI before and after? I'm guessing the univariate is low because the mean doesn't mean much...}

%\begin{figure}[h!]
%	\centering
%	\includegraphics[scale=0.3]{results/May_Pr_3}
%    \caption{Prior Drift of 3 attributes with Brightness as one of the attributes}
%    \label{fig:Prior3}
%\end{figure}
\begin{figure}
	\centering
	\includegraphics[scale=0.3]{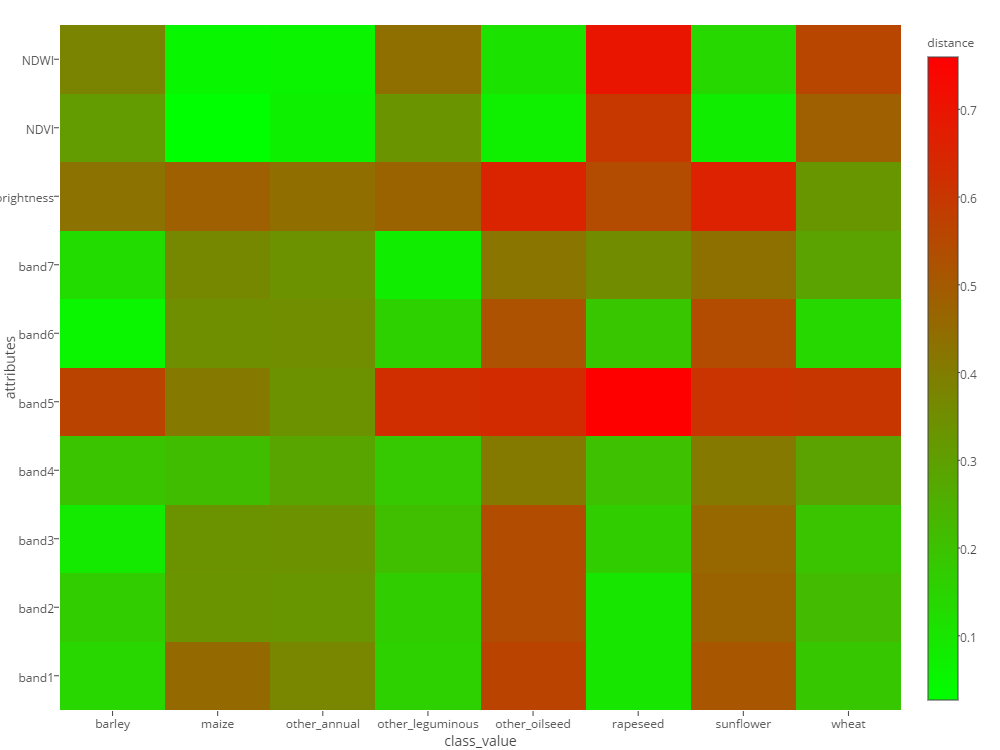}
    \caption{Conditioned Covariate Drift of individual attributes}
    \label{fig:CC1}
\end{figure}

Figure \ref{fig:CC1} gives a heat map of the drift of each covariate conditioned on the class.  The x-axis gives the classes and the y-axis gives each of the covariates. This illustrates how drift can vary greatly from class to class. For \texttt{wheat} and \texttt{rapeseed/canola}, NDVI and NDWI are changing substantially between May and November, which is explained by the fact that May is the peak season for these winter crops while they have been harvested in November. As a result there are significantly changes in the NDVI --~which is a proxy for plant health~-- and NDWI --~which is a proxy for the water content of the leaves. Interestingly, \texttt{maize/corn} doesn't drift for NDVI and NDWI as these crops are growing after May and harvested before November; they thus keep the same ``bare soil'' reflectance. % For \texttt{other annual} none of the attributes drifts greatly between the two time periods.  For \texttt{barley} and \texttt{other leguminous} there is substantial drift in \texttt{band5} but little drift in the other bands.  In contrast \texttt{other oilseed} and \texttt{sunflower} exhibit substantial drift across all bands.

\begin{figure}
	\centering
	\includegraphics[scale=0.6]{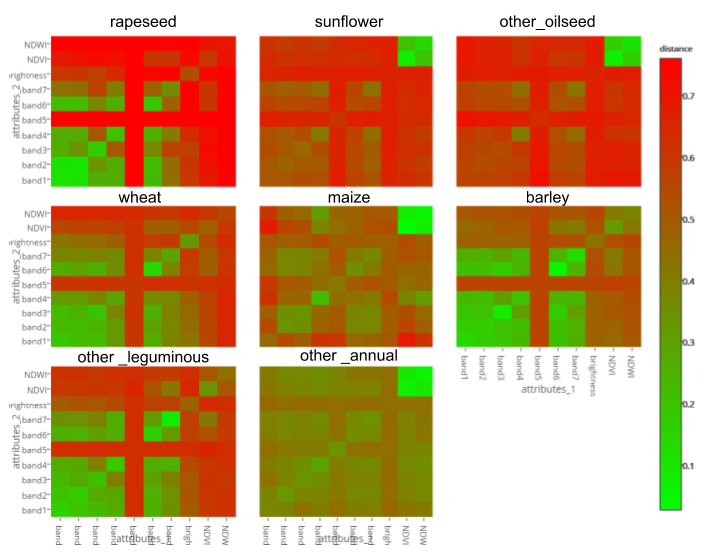}
    \caption{Conditioned Covariate Drift for pairs of attributes}
    \label{fig:CC2}
\end{figure}
Figure \ref{fig:CC2} provides heat maps for each pair of attributes conditioned by each class.  It~also illustrates the monotonicity of drift magnitude. The drift for any pair of attributes given a class must always be at least as high as the univariate drift of either of the attributes given that class.
For instance, from Figure \ref{fig:CC1}, we can observe the attributes \texttt{band4} and \texttt{band7} 
drift the lowest among the other \texttt{band} attributes given the classes \texttt{other\_oilseed} and \texttt{sunflower}.
This translates to a low joint drift magnitude under the same classes in Figure~\ref{fig:CC2}.

\begin{figure}
	\centering
	\includegraphics[scale=0.3]{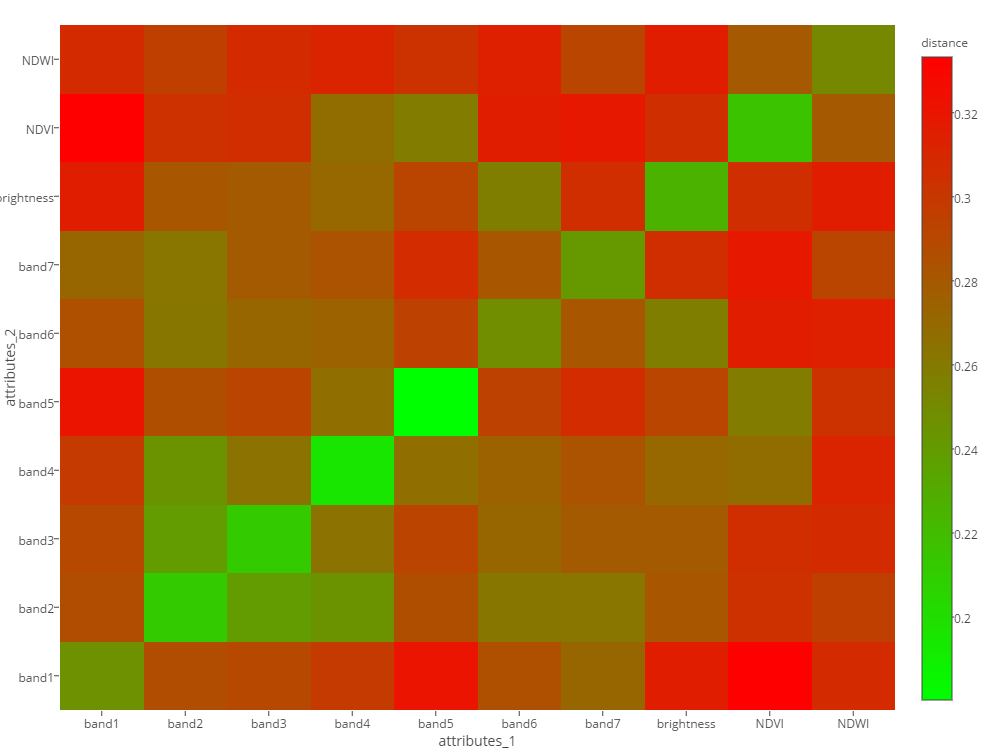}
    \caption{Posterior Drift conditioned on pairs of attributes}
    \label{fig:Pos2}
\end{figure}
Figure \ref{fig:Pos2} shows the posterior drift conditioned on pairs of attributes.  It might at first sight seem anomalous that there should be posterior drift of up to 0.34 when the class is conditioned on specific pairs of attributes, but no posterior drift when the class is considered in isolation.
As explained above, this arises because the only way in which $P(X)$ can change while $P(Y)$ remains invariant is for $P(Y\mid X)$ to change.  It is particularly revealing that the posterior drift within each individual x-value is low, while for some combinations of x-values it becomes relatively high.  This demonstrates the value of evaluating the drift across different combinations of attributes. We find here again high values for NDVI, NDWI and band 5, which is explained by the difference in the agricultural season.

%\section{Methods}
%\input{"methods/methods.tex"}
%\subsection{Frequency Matrix}
%\input{"methods/frequency_matrix.tex"}
%\input{"methods/frequency_map.tex"}
%\subsection{Obtaining Probability Distributions}
%\input{"methods/probabilities.tex"}
%\subsection{Drift Distance}
%\input{"methods/drift_distance.tex"}
%\subsection{Results Format}
%\input{"methods/results.tex"}
%\section{Results}
%\input{"results/results.tex"}

%\section{Discussion}\label{sec:discussion}

\section{Conclusions and Future Research}\label{sec:conclusion}

\sloppy Concept drift is in some senses the great elephant in the room for machine learning. The~world is continually changing, but we have a dearth of techniques for understanding the nature of these changes as they apply to specific machine learning contexts.  %In practice this problem appears to be addressed by learned models being decommissioned on a regular schedule and replaced by models learned on more recent data.  However, without an understanding of the nature of the drift with which an application is faced, how can one optimize such a schedule or the windows from which models are learned?
We have a growing body of sophisticated methods for learning in the context of concept drift \citep{Gaber05,Gama09,Aggarwal09,LearningUnderConceptDrift,Bifet2011PAKDD,Nguyen2014, Krempl2014survey,SurveyConceptDriftAdaptation}. There is a need to develop a supporting body of techniques for understanding the phenomena that these methods address and thereby understanding the relative capabilities of these methods in the face of different expressions of that phenomena.

The current work extends our investigations of how drift may be analyzed and described.  We have previously argued that it is important to augment with quantitative descriptions the qualitative descriptions of drift which were the previous state-of-the-art  \citep{webb2015characterizing}.  In the current work we argue that it is important to augment global summaries of drift with detailed analyses of drift within the marginal distributions.

There are two reasons for doing so.  The first is that drift within the marginal distributions provides greater detail, such as which variables and combinations of variables are most and least subject to drift.  The second reason is that as dimensionality increases the global drift magnitude must also increase, and hence, with high dimensionality, drift magnitude will tend toward 1.0 and cease to be informative.

These preliminary techniques for mapping concept drift leave substantial scope for refinement.
\begin{itemize}
\item It may prove useful to handle numeric data directly without requiring discretization.
\item In the current work we use maximum likelihood estimates of the probability distributions. These are likely to be imprecise, adding noise to the estimates which will accumulate as dimensionality increases. Methods to address this issue are likely to be important when seeking to map high-dimensional data.
\item  For very high dimensional data it will not be feasible to present and consider every pairwise marginal distribution.
There is a need for techniques either to identify and highlight the marginals in which the drift is most interesting, or to allow a user to explore the space of marginals in an effective manner.
\item In the airlines example,  inter-day and inter-week drift demonstrated very different patterns, each of which was revealing of different dynamics in the data. This well illustrates the importance of identifying informative granularity for analysis.  In some domains this may be readily apparent to the relevant experts.  However, there are likely to be domains where the analyst does not have access to such expertise and it would be useful to have tools to automatically identify appropriate granularities for analysis.
\end{itemize}%\todo{GW: It would be good to add further directions for future research. Any suggestions?}

We have presented practical techniques for creating and communicating the nature of drift affecting specific applications. Our case studies on three real-world datasets demonstrate that these techniques can reveal insights into the nature of specific instances of drift that cannot be obtained by any prior method.  We hope that these techniques will have practical application in addressing this very real and present problem. As a service to the community we are in the process of establishing an online server to which users can upload data to be analysed by our tools.  In the interests of reproducible research we make the software necessary to reproduce our results available at \url{https://github.com/LeeLoongKuan/DriftMapper} and \url{https://github.com/LeeLoongKuan/DataAnalysisR}. The first of these produces the drift maps in numeric form while the second creates the heat map and line plot visualizations.

\section*{Acknowledgements}

This work was supported by the Australian Research Council under awards DP140100087 and DE170100037. This material is based upon work supported by the Air Force Office of Scientific Research, Asian Office of Aerospace Research and Development (AOARD) under award number FA2386-15-1-4007.

The authors would like to thank the colleagues from CESBIO (Jordi Inglada, Arthur Vincent, Marcela Arias, Benjamin Tardy, David Morin and Isabel Rodes) for
providing the Satellite dataset (data and labels).

\appendix

\section{Proof that drift magnitude is monotone under increasing dimensionality}\label{appendix:monotonicity}

We here prove that Total Variation Distance is monotone under increasing dimensionality. The proof generalizes trivially to Hellinger Distance.  Note that where one set of variables is conditioned on another, it is the dimensionality of the conditioned variable rather than the conditioning variables over which this monotone increase in distance applies.

Let $X,Z$ be sets of covariates.
\begin{eqnarray*}
\tvd_{t,u}(X) & \leq & \tvd_{t,u}(X,Z) \\
& \Updownarrow & \\
\frac{1}{2}\sum_{\scriptsize\begin{array}{c}
{\bar{x}\in\dom(X)}
\end{array}}\left|P_t(\bar{x})-P_u(\bar{x})\right| & \leq & \frac{1}{2}\sum_{\scriptsize\begin{array}{c}
{\bar{x}\in\dom(X)} \\
{\bar{z}\in\dom(Z)}
\end{array}}\left|P_t(\bar{x},\bar{z})-P_u(\bar{x},\bar{z})\right| \\
& \Updownarrow & \\
% \sum_{\scriptsize\begin{array}{c}
% {\bar{x}\in\dom(X)}
% \end{array}}\left|P_t(\bar{x})-P_u(\bar{x})\right| & \leq & \sum_{\scriptsize\begin{array}{c}
% {\bar{x}\in\dom(X)} \\
% {\bar{z}\in\dom(Z)}
% \end{array}}\left|P_t(\bar{x},\bar{z})-P_u(\bar{x},\bar{z})\right| \\
% & \Updownarrow & \\
\sum_{\scriptsize\begin{array}{c}
{\bar{x}\in\dom(X)}
\end{array}}\left|\sum_{\scriptsize\begin{array}{c}
{\bar{z}\in\dom(Z)}
\end{array}} P_t(\bar{x},\bar{z})- \sum_{\scriptsize\begin{array}{c}
{\bar{z}\in\dom(Z)}
\end{array}} P_u(\bar{x},\bar{z})\right| & \leq & \sum_{\scriptsize\begin{array}{c}
{\bar{x}\in\dom(X)}
\end{array}}\sum_{\scriptsize\begin{array}{c}
{\bar{z}\in\dom(Z)}
\end{array}}\left|P_t(\bar{x},\bar{z})-P_u(\bar{x},\bar{z})\right| \\
& \Updownarrow & \\
\sum_{\scriptsize\begin{array}{c}
{\bar{x}\in\dom(X)}
\end{array}}\left|\sum_{\scriptsize\begin{array}{c}
{\bar{z}\in\dom(Z)}
\end{array}} P_t(\bar{x},\bar{z})-P_u(\bar{x},\bar{z})\right| & \leq & \sum_{\scriptsize\begin{array}{c}
{\bar{x}\in\dom(X)}
\end{array}}\sum_{\scriptsize\begin{array}{c}
{\bar{z}\in\dom(Z)}
\end{array}}\left|P_t(\bar{x},\bar{z})-P_u(\bar{x},\bar{z})\right|
\end{eqnarray*} 

\bibliographystyle{spbasic}
\bibliography{bibliography}

\end{document}